\documentclass[11pt,letterpaper]{article}
\usepackage{emnlp2016}
\usepackage{times}
\usepackage{latexsym}
\usepackage{tabulary}
\usepackage{float}
\usepackage{bbding}
\usepackage{tablefootnote}

\emnlpfinalcopy

\title{Overview for the Second Shared Task on Language Identification in Code-Switched Data}

\author{Giovanni Molina, Nicolas Rey-Villamizar, Thamar Solorio\\
Dept. of Computer Science \\
University of Houston \\
Houston TX, 77004 \\
{\tt \{gemolinaramos,nrey\}@uh.edu solorio@cs.uh.edu}
\\
\AND Fahad AlGhamdi, Mahmoud Ghoneim, Abdelati Hawwari, Mona Diab\\
Dept. of Computer Science \\
George Washington University \\
Washington DC, 20052 \\
{\tt \{fghamdi,mghoneim,abhawwari,mtdiab\}@gwu.edu}}

\date{}

\begin{document}

\maketitle

\begin{abstract}
We present an overview of the second shared task on language identification in code-switched data. For the shared task, we had code-switched data from two different language pairs: Modern Standard Arabic-Dialectal Arabic (MSA-DA) and Spanish-English (SPA-ENG). We had a total of nine participating teams, with all teams submitting a system for SPA-ENG and four submitting for MSA-DA. Through evaluation, we found that once again language identification is more difficult for the language pair that is more closely related. We also found that this year's systems performed better overall than the systems from the previous shared task indicating overall progress in the state of the art for this task.
  
\end{abstract}

\section{Introduction}
With the First Shared Task on Language Identification in Code-Switched Data we managed to raise awareness and shine a spotlight on the difficult problem of automatic processing of Code-Switched (CS) text. This year our goal is not only to maintain research interest in the problem, but also to bring in new ideas to tackle it. With the continuing growth of social media usage, it is more likely to find CS text and thus the problem becomes more relevant. 

Code-switching is a linguistic phenomenon where two or more languages are used interchangeably in either spoken or written form. It is important to study and understand CS in text because any advancement in solving the problem will positively contribute to other NLP tasks such as Part-of-Speech tagging, parsing, machine translation, among others. In order to achieve this, we organized this year's shared task with the intention of providing our peers with new annotated data, to further develop a universal annotation scheme for CS text and most significantly to motivate high quality research. 


\begin{table}[h]
\centering
\renewcommand{\arraystretch}{1}
\setlength{\tabcolsep}{0.4ex}
\small
\begin{tabular}{|c|p{5.5cm}|}
\hline
{\bf Language Pair} & {\bf Example}\\\hline
MSA-DA & \textbf{Buckwalter:\tablefootnote{We use the Buckwalter encoding to present all the Arabic data in this paper: It is an ASCII only transliteration scheme, representing Arabic orthography strictly one-to-one.}}  *hbt AlY AlmHAkm wAlnyAbAt fy Ehd mbArk HwAly 17mrp, EAyzyn nqflhm rqm mHtrm, xmsyn mvlA 

\textbf{English Translation:} \textit{I went to courts in Mubark's era almost 17 times. I would like to reach a respectful number, for example 50 times.}\\\hline

SPA-ENG & \textbf{Original:} Styling day trabajando con @username vestuario para \#ElFactorX y soy hoy chofer. I  will get you there in pieces ☔⚡im a Safe Driver. \\
& \textbf{English Translation:} \textit{Styling day working with @username wardrobe for \#ElFactorX and today I drive. I will get you there in pieces ☔⚡im a Safe Driver.} \\\hline
\end{tabular}

\caption{Twitter code-switched data examples.}\label{tab:cs_examples}

\end{table}
\addtocounter{footnote}{1}
This shared task covers two different language pairs and is focused on social media data obtained from Twitter. The language pairs used this time are Spanish-English (SPA-ENG) and Modern Standard Arabic-Dialect Arabic (MSA-DA). These languages are widely used around the world and are good examples of language pairs that are easily interchanged by the speakers. Participants are tasked with predicting the correct language label for each token in the unseen test sets. 

We provide a full description of the shared task in the following section and we talk about related work in section 3. The data sets used for the task are described in section 4, followed by an overview of the submitted systems in section 5. Finally, we show the results, lessons learned and conclusion in sections 6, 7 and 8, respectively.

\section{Task Description}
Similar to the first edition, the task consists of labeling each token/word in the input test data with one out of 8 labels: \emph{lang1, lang2, fw, mixed, unk, ambiguous, other} and \emph{named entities (ne)}. The labels \emph{fw } and \emph{unk} were added in this edition of the shared task and are derived from tokens that used to be labeled as \emph{other}. The \emph{lang1, lang2} labels correspond to the two languages in the language pair, for example with SPA-ENG, \emph{lang1} would be ENG and \emph{lang2} would be SPA. The \emph{fw} label is used to tag tokens that belong to a language other than the two languages in the language pair. The \emph{mixed} label is used to tag words composed of code-switched morphemes, such as the word uber\textit{eando} ('driving for Uber') in SPA-ENG. The \emph{unk} label is used to tag tokens that are gibberish or unintelligible. The \emph{ambiguous} label is used to tag words that could be labeled as either language in the language pair and the context surrounding the word is not enough to determine a specific language, for example the word \emph{a} is a determiner in English and a preposition in Spanish and it can be hard to tell which language it belongs to without the surrounding context. The \emph{other} label is used to tag usernames, emoticons, symbols, punctuation marks, and other similar tokens that do not represent words. Lastly, the \emph{ne} label is used to tag named entities, which are proper nouns and must be identified correctly in order to properly conduct an analysis of CS data. This is due to the fact that named entities are usually kept the same even as languages switch. Named entities are problematic even for human annotators and require a lot of work, including defining absolute and correct guidelines for annotation.

In Table~\ref{tab:cs_examples} we show examples of code-switched tweets that are found in our data. We have posted the annotation guidelines for SPA-ENG, but it can be generalized to the MSA-DA language pair as well. This is possible because we want to have a universal set of annotation labels that can be used to correctly annotate new data with the least amount of error possible. We keep improving the guidelines to accommodate findings from the previous shared task as well as new relevant research.

\section{Related Work}
The earliest work on CS data within the NLP community dates back to research done by Joshi \shortcite{joshi:82} on an approach to parsing CS data. Following work has been described in the First Shared Task on Language Identification in Code-Switched Data held at EMNLP 2014 ~\cite{solorio:14}. Since the first edition of the task, new research has come to light involving CS data. 

There has been work on language identification of different language pairs in CS text, such as improvements on dialect identification in Arabic \cite{al2015aida2} and detection of intra-word CS in Dutch and dialect varieties \cite{nguyen122016automatic}. There has also been work on POS tagging and parsing such as parsing of bilingual code-switched text ~\cite{vilares:15b}, POS tagging of Hindi-English CS social media text ~\cite{sequierapos,jamatia2015part} and shallow parsing of Hindi-English CS social media text \cite{sharma-EtAl:2016:N16-1}. Another area where there has been some new research work is in sentiment analysis, such as emotion detection in Chinese-English code-switched texts \cite{lee:15} and sentiment analysis on Spanish-English Twitter posts \cite{vilares:15a}. 

\cite{kosoff} carried out a sociolinguistic investigation focused on the use of code-switching in the complex speech community of Egyptian Twitter users. It studies the combinations of Modern Standard Arabic(MSA), Egyptian Colloquial Arabic, English, and Arabizi; whether it is a Modern Standard Arabic or Egyptian Colloquial Arabic. The research goal was to describe the code switching phenomena situation found in this Egyptian twitter community.

We also had contributions of new CS corpora, such as a collection of Arabic-Moroccan Darija social media CS data \cite{SAMIH16.341}, a collection of Turkish-German CS tweets \cite{ETINOLU16.1151}, a large collection of Modern Standard Arabic and Egyptian Dialectal Arabic CS data \cite{DIAB16.1161} and a collection of sentiment annotated Spanish-English tweets \cite{VILARES16.43}. Other work includes improving word alignment and MT models using CS data \cite{huang2014improving}, improving OCR in historical documents that contain code-switched text \cite{garrette:15}, the definition of an objective measure of corpus level complexity of code-switched texts \cite{GAMBCK16.669} and \cite{BEGUM16.932} presented an annotation scheme for annotating the pragmatic functions of CS in Hindi-English code-switched tweets. There is still more research to be done involving CS data, and we hope this second edition of the shared task will help motivate further research.

\section{Data Sets}
The data for the shared task was collected from Twitter. We decided to use Twitter because it has a large user base from a multitude of countries and this provides a good space to find code-switched data. Twitter also provides an API which makes it easier to crawl and collect data. However, there are limitations to the amount of data that can be collected and restrictions to how we can share  data.

\begin{table}[H]
\centering
\renewcommand{\arraystretch}{1}
\setlength{\tabcolsep}{0.3ex}
\small
\begin{tabular}{|c|c|c|c|}
\hline
{\bf Language Pair} & {\bf Training} & {\bf Development} & {\bf Test} \\\hline
MSA-DA & 8,862 & 1,117 & 1,262 (1,258) \\
SPA-ENG & 8,733 & 1,857 & 18,237 (10,716) \\\hline
\end{tabular}
\caption{Data set statistics for SPA-ENG and MSA-DA.}\label{tab:data set_stats}
\end{table}

In Table~\ref{tab:data set_stats} we show the statistics for the data sets used in the shared task. These statistics were taken at the moment the data was released, but they can change with time due to tweets becoming unavailable. To account for this limitation, we used the maximum amount of tweets that all participants had in common after submission of the test data, which we show in parenthesis.

\subsection{SPA-ENG}
For SPA-ENG, we used the training and test corpora from the EMNLP 2014 shared task as this year's training corpus and development corpus, respectively. However, the previous shared task did not have the same labels we are using this year. We had in-lab annotators follow a simple mapping process: they went through all the tokens previously labeled as \emph{other} that are not usernames, emoticons or symbols and either changed them to \emph{fw}, \emph{unk} or kept them as \emph{other}. The annotators used the annotation guidelines posted on the shared task website to make the decisions required for the mapping.

We made improvements to the quality of the data with the help of our in-lab annotators. We performed quality checks for label inconsistencies by manually looking at the labels of word types that had more than one label and changing them to the correct one if needed. We also verified and fixed the tokenization of the data to remove single whitespaces that appeared as tokens and to correctly tokenize emoticons and other symbols.

Building the new test corpus consisted of first finding code-switched data and secondly annotating the data. Finding code-switched tweets is not an easy task, so we followed a series of steps to help us locate a good amount of code-switched tweets. First, similar to the previous task, we selected geographical areas where there is a strong presence of bilingual speakers, specifically Miami and New York. Secondly, we performed a search within these areas for popular Spanish language radio stations with active Twitter accounts. Then, we selected the Twitter accounts of radio stations that code-switched in their tweets and collected them. From here, we looked for code-switched tweets within the accounts of users that are followed by and that follow the radio stations and collected them as well. Finally, we examined the accounts of users that interact with these users and check for code-switched tweets. In total, we obtained tweets from 21 users and a total of 61,943 tweets.

The annotation process consisted of three steps: in-lab pre-annotation, crowd-sourcing and in-lab quality check. Our in-lab pre-annotation was performed by training an SVM with the training corpus from the previous shared task and then annotating our test data with the language label. We then used a pattern matching algorithm with regular expressions to label tokens with \emph{other}. Following that, we ranked our word types by token frequency score and selected the top thousand most frequent words and manually verified the labels and fixed them where necessary. We then propagated this to the entire data set and separated the affected tokens and considered them as annotated. Next, we took the rest of the unannotated data set and used CrowdFlower to annotate it. We used a subset of the training corpus (one that contained roughly equally distributed tokens for each label) as our gold data for quality control within the crowd-sourcing task. After the first round of annotation was completed, we took tokens with a confidence score under 0.75 and resubmitted them to CrowdFlower in an attempt to improve the quality of the tags. Finally, we manually went through the most frequent tokens that had more than one label assigned and verified them to get rid of inconsistencies and further improve the quality. This whole process of annotation cost us roughly \$1,100, which comes out to about \$0.02 per tweet.  
In order to have the best possible quality in the data, we first selected the top 35K tweets ranked by overall tweet confidence, which was taken to be the lowest confidence among the tokens for that tweet. From here we selected all the code-switched tweets and 10k monolingual tweets. This became the official test data set for the shared task. We later decided to remove tweets that contain URLs in them for consistency with the training and development corpora as they did not contain URLs. This is the subset of test data that we end up using to rank the participating systems.

\begin{table}
\centering
\renewcommand{\arraystretch}{1}
\setlength{\tabcolsep}{0.5ex}
\small
\begin{tabular}{|c|c|}
\hline
{\bf Monolingual Tweets} & {\bf Code-Switched Tweets} \\\hline
4,626 & 6,090 \\\hline
{\bf Label} & {\bf Tokens} \\\hline
ambiguous & 4 \\
lang1 & 16,944 \\
lang2 & 77,047 \\
mixed & 4 \\
ne & 2,092 \\
fw & 19 \\
other & 25,311 \\
unk & 25 \\
{\bf Total} & {\bf 121,446} \\\hline
\end{tabular}
\caption{Test Data statistics for SPA-ENG.}\label{tab:test_stats}
\end{table}

In Table~\ref{tab:test_stats} we show the statistics of the data set used to evaluate the participating systems. This includes only the tweets that all the participants managed to crawl from Twitter, and it is not the complete data set.

\subsection{MSA-DA}
For the MSA-DA language pair, the Egyptian dialect is used as the Arabic dialect, EGY. We combined the Train, Test-1, and Test-2 corpora that we used in the EMNLP 2014 shared task to create the new training and development corpora. The data was crawled and collected from Twitter. We perform a number of quality checks on the old data to overcome any issues that the participants may face. One of these checks is that all the old tweets are re-crawled from Twitter to reduce the percentage of missing tweets. The missing tweets and the tweets that contained white spaces were removed. This step was performed as a validation step. After the validation step, we accepted and published 9,979 Tweets (8,862 tweets for the training set, and 1,117 tweets for the development set).

Building a new test corpus required crawling new data and annotating the crawled data. As we did in the previous shared task, we used the “Tweepy” library to harvest the timeline of 26 Egyptian Public Figures. We have some filtration criteria that we applied on the new test set. Since we are using the tweets that we introduced in the EMNLP 2014 CS shared task, we set the crawling script to harvest only the tweets that were created in 2014, to maintain consistency in topics with the training/dev data sets. Also, the tweets that contain URLs and re-tweets were excluded. The total number of harvested tweets after applying the filtration criteria was 12,028 tweets. This number of tweets was bigger than what we needed for the test set. So, we chose only 1,262 tweets. However, before choosing the 1,262 tweets, we wanted to consider the public figure whose tweets contain more code-switching points. So, we input all the tweets into the Automatic Identification of Dialectal Arabic (AIDA2) tool \cite{al2015aida2} to perform token level language identification for the EGY and MSA tokens in context. According to AIDA2’s output we chose a certain percentage of tweets from the Public Figure whose CS percentage in his/her tweets is more than 35\%. We used the improved version of the “Arabic Tweets Token Assigner” which is made available through the shared task website \footnote{http://care4lang1.seas.gwu.edu/cs2/call.html} to avoid the misalignment issues and guarantee consistency. 

Two Egyptian native speakers were asked to perform the annotation. They were trained to get involved with the project's concepts and related linguistic issues to increase their productivity and accuracy.  Our Annotation team used two types of CS tag-sets: a) a rich version which is Arabic dialects oriented and it is used in our lab; and, b) a reduced CS tag-set which is consistent. The two tag-sets are mappable to each other (we mapped our tag set to the six tags). In our annotation, we used \textit{lang1} to represent MSA, and \textit{lang2} for Egyptian words, \textit{ambiguous} when the context can’t help  decide if a word is MSA or DA, and foreign word (\textit{fw}) for non-Arabic word even it is in Arabic script or Latin script. To manage the annotation quality, the annotations were checked when initially performed and then checked again at the end of the task. The Inter-Annotator Agreement (IAA) was measured by using 10\% of the total number of data to ensure the performance and agreement among annotators. A specialist linguist carried out adjudication and revisions of accuracy measurements. We approached a stable Inter Annotator Agreement (IAA) of over 90\% pairwise agreement.

\begin{table}[H]
\centering
\renewcommand{\arraystretch}{1}
\setlength{\tabcolsep}{0.5ex}
\small
\begin{tabular}{|c|c|}
\hline
{\bf Monolingual Tweets} & {\bf Code-Switched Tweets} \\\hline
1,044 & 214 \\\hline
{\bf Label} & {\bf Tokens} \\\hline
ambiguous & 117 \\
lang1 & 5,804 \\
lang2 & 9,630 \\
mixed & 1 \\
ne & 2,363 \\
fw & 0 \\
other & 2,743 \\
unk & 0 \\
{\bf Total} & {\bf 20,658 } \\\hline
\end{tabular}
\caption{Test Data statistics for MSA-DA.}\label{tab:MSA_DA_test_stats}
\end{table}
Table~\ref{tab:MSA_DA_test_stats} shows the statistics of the MSA-DA test set used to evaluate the participating systems. It contains only the tweets that all the participants managed to crawl from Twitter, and it's not the complete data set.


\begin{table*}[t]
\centering
\renewcommand{\arraystretch}{1}
\setlength{\tabcolsep}{1ex}
\small
\begin{tabular}{|m{3cm}| m{1.5cm} m{1.5cm} c m{3cm} c c c c|}
\hline
{\bf System} & {\bf Traditional Machine Learning} & {\bf Deep Learning} & {\bf Rules} & {\bf External Resources} & {\bf LM} & {\bf Case} & {\bf Affixes} & {\bf Context} \\\hline
\cite{gwu:16} & CRF & - & - & SPLIT, Gigaword & \Checkmark & - & - & - \\\hline
\cite{mcgill:16} & CRF & - & - & fastText & - & \Checkmark & \Checkmark & $\pm 1$ \\\hline
\cite{uwgroup:16} & - & CNN, LSTM & - & - & - & - & - & - \\\hline
\cite{howard:16} & Logistic Regression & - & - & GNU Aspell, NER, POS tagger  & - & - & \Checkmark & - \\\hline
\cite{columbia:16} & - & - & \Checkmark & NER, Dictionaries & - & - & - & $\pm 1 $ \\\hline
\cite{hhu:16} & CRF & LSTM & - & Gigaword, word2vec & - & \Checkmark & \Checkmark & $\pm 1$ \\\hline
\cite{nepswitch:16} & CRF & - & - & - & - & \Checkmark & \Checkmark & -\\\hline
\cite{ntnu:16} & CRF & - & - & Babelnet, Babelfy & - & \Checkmark & \Checkmark & $\pm 2$ \\\hline
\end{tabular}
\caption{Summary of the architectures of the systems submitted. }\label{tab:systems}
\end{table*}
\section{Survey of Shared Task Systems}
This year we received submissions from nine different teams, which is two more teams than the previous shared task. All teams participated in the SPA-ENG task, while four teams also participated in the MSA-DA task. There was a wide variety of system architectures ranging from simple rule based systems all the way to more complex machine learning implementations. Most of the systems submitted did not change anything in the implementation to tackle one language pair or the other, which implies that the participants were highly interested in building language independent systems that could be easily scaled to multiple language pairs. 

In Table \ref{tab:systems} we show a summary of the the architectures of the systems submitted by the participants. All teams, with the exception of \cite{columbia:16}, used some sort of machine learning algorithm in their systems. The algorithm of choice by most participants was the Conditional Random Fields (CRF). This is no surprise since CRFs fit the problem nicely due to the sequence labeling nature of the task as it was evidenced in the high performance by CRFs achieved in the previous shared task. 

A new addition this year is the use of deep learning algorithms by two of the participants. Deep learning is now much more prevalent in NLP than it was two years ago when the previous shared task was held. \cite{uwgroup:16} used a convolutional neural network (CNN) to obtain word vectors which are then fed as a sequence to a bidirectional long short term memory recurrent neural network (LSTM) to map the sequence to a label. The system submitted by \cite{hhu:16} used the output of a pair of LSTMs along with a CRF and post-processing to obtain the final label mapping. These systems are perhaps more complex than traditional machine learning algorithms, but the trade off for performance is evident in the results.

Most of the participants included some sort of external resource in their system. Among them we can find large monolingual corpora, language specific dictionaries, Part-of-Speech taggers, word embeddings and Named Entity Recognizers. Other features used in some of the systems were language models, word case information (Title, Uppercase, Lowercase), affixes and surrounding context. 

\begin{table*}[t]
\centering
\renewcommand{\arraystretch}{1}
\setlength{\tabcolsep}{0.5ex}
\small
\begin{tabular}{|c|c|c|c|c|}
\hline
{\bf Test Set} & {\bf System} & {\bf Monolingual F-1} & {\bf Code-switched F-1} & {\bf Weighted F-1}  \\\hline
& Baseline & 0.54 & 0.69 & 0.607 \\
& \cite{gwu:16}*$^\dagger$ & 0.83 & 0.69 & 0.77 \\
& \cite{columbia:16}$^\delta$ & 0.83 & 0.75 & 0.79  \\
& \cite{mcgill:16} & 0.86 & 0.79 & 0.83 \\
& \cite{nepswitch:16} & 0.90 & 0.86 & 0.88 \\
SPA-ENG & \cite{ntnu:16} & 0.91 & 0.87 & 0.89 \\
& IIIT Hyderabad$^\top$ & 0.91 & 0.88 & 0.898 \\
& \cite{uwgroup:16} & 0.91 & 0.88 & 0.898 \\
& \cite{hhu:16}* & 0.92 & 0.88 & 0.90 \\
& \cite{howard:16} & 0.93 & 0.9 & \textbf{0.913} \\\hline

& Baseline & 0.47 & 0.31 & 0.44 \\
& \cite{nepswitch:16} & 0.72 & 0.34 & 0.66  \\
& \cite{gwu:16}-1* & 0.75 & 0.43 & 0.69 \\
MSA-DA & \cite{uwgroup:16} & 0.83 & 0.25 & 0.73  \\
& \cite{gwu:16}-2* & 0.83 & 0.37 & 0.75  \\
& \cite{hhu:16}* & 0.89 & 0.50 & \textbf{0.83} \\\hline
\end{tabular}
\caption{Tweet level performance results. We ranked the systems using the weighted average F-measure, Weighted-F1. A '$^\dagger$' denotes a late submission. A '*' denotes systems submitted by co-organizers of the shared task. A '$^\delta$' denotes the participant submission is missing a small number of tokens from one tweet. A $^\top$ denotes the participant did not submit a system description.}\label{tab:res_tweet}
\end{table*}

\begin{table*}[t]
\centering
\renewcommand{\arraystretch}{1}
\setlength{\tabcolsep}{0.2ex}
\small
\begin{tabular}{|c|c|c|c|c|c|c|c|c|c|c|c|c|}
\hline
{\bf Test Set} & {\bf System} & {\bf lang1} & {\bf lang2} & {\bf NE} & {\bf other} & {\bf ambiguous} & {\bf mixed} & {\bf fw} & {\bf unk} & {\bf Avg-F} & {\bf Avg-A} \\\hline
& \cite{columbia:16} & 0.478 & 0.689 & 0.153 & 0.466 & 0.0 & 0.0 & 0.01 & 0.003 & 0.603 & 0.536 \\
& Baseline & 0.595 & 0.852 & 0.0 & 0.979 & 0.0 & 0.0 & 0.0 & 0.0 & 0.828 & 0.811 \\
& \cite{gwu:16}*$^\dagger$ & 0.828 & 0.959 & 0.256 & 0.982 & 0.0 & 0.0 & 0.0 & 0.0 & 0.933 & 0.938 \\
& \cite{mcgill:16} & 0.873 & 0.965 & 0.379 & 0.993 & 0.0 & 0.0 & 0.0 & 0.0 & 0.947 & 0.949 \\
& \cite{nepswitch:16} & 0.919 & 0.978 & 0.481 & 0.994 & 0.0 & 0.0 & 0.0 & 0.0 & 0.964 & 0.965 \\
SPA-ENG & \cite{ntnu:16} & 0.928 & 0.979 & 0.510 & 0.996 & 0.0 & 0.0 & 0.0 & 0.045& 0.967 & 0.965 \\
& \cite{uwgroup:16} & 0.929 & 0.982 & 0.480 & 0.994 & 0.0 & 0.0 & 0.0 & 0.0 & 0.968 & 0.969 \\
& \cite{hhu:16}* & 0.930 & 0.980 & 0.551 & 0.995 & 0.0 & 0.0 & 0.0 & 0.034 & 0.968 & 0.967 \\
& IIIT Hyderabad$^\top$ & 0.931 & 0.979 & 0.645 & 0.991 & 0.0 & 0.0 & 0.0 & 0.013& 0.969 & 0.966 \\
& \cite{howard:16} & 0.938 & 0.984 & 0.603 & 0.996 & 0.0 & 0.0 & 0.0 & 0.029 & \textbf{0.973} & 0.973 \\\hline

& Baseline & 0.534 & 0.421 & 0.0 & 0.883 & 0.0 & 0.0 & - & - & 0.463 & 0.513 \\
& \cite{uwgroup:16} & 0.603 & 0.603 & 0.468 & 0.712 & 0.0 & 0.0 & - & - & 0.594 & 0.599 \\
& \cite{nepswitch:16} & 0.699 & 0.722 & 0.745 & 0.975 & 0.0 & 0.0 & - & - & 0.747 & 0.747 \\
MSA-DA & \cite{gwu:16}-2* & 0.767 & 0.833 & 0.828 & 0.986 & 0.0 & 0.0 & - & - & 0.828 & 0.826 \\
& \cite{gwu:16}-1* & 0.802 & 0.860 & 0.827 & 0.988 & 0.0 & 0.0 & - & - & 0.851 & 0.852 \\
& \cite{hhu:16}* & 0.854 & 0.904 & 0.77 & 0.957 & 0.0 & 0.0 & - & - & \textbf{0.876} & 0.879 \\\hline
\end{tabular}
\caption{Token level performance results. We ranked the systems using the weighted average F-measure, Avg-F. A '-' indicates that there were no tokens labeled under this class in the test data set. A '$^\dagger$' denotes a late submission. A '*' denotes systems submitted by co-organizers of the shared task. A '$^\delta$' denotes the participant submission is missing a small number of tokens from one tweet. A $^\top$ denotes the participant did not submit a system description.}\label{tab:res_token}
\end{table*}

\section{Results}
Same as the previous shared task, we used the following metrics to evaluate the submitted systems: Accuracy, Precision, Recall and F-measure. We use regular F-measure to rank the systems at the tweet level and the weighted average F-measure to rank the systems at token level to account for the imbalanced distribution of the labels.

To evaluate the systems, we first took the subset of tweets that all participants had in common to provide a fair comparison among them. We designed a simple lexicon-based baseline system by taking only the lexicon for \emph{lang1, lang2} from the training corpus. We labeled symbols, emoticons, usernames, punctuation marks and URLs as other. If we find an unseen token or have a tie, we assign the majority class label. We compare the results of all participants to this baseline.

To calculate the performance of the systems at the tweet level, we use the predicted token level information to determine if a tweet is code-switched or monolingual. If the tweet has at least one token from each language (\emph{lang1, lang2}), then it is labeled a code-switched. Otherwise, the tweet is labeled as monolingual. Table \ref{tab:res_tweet} shows the tweet level results for all submitted systems in both language pairs, ranked by the average weighted f-measure. We can see that the best performing systems in SPA-ENG perform better than the best performing systems in MSA-DA, which indicates that this is a more difficult task for the MSA-DA language pair as both languages are closely related, as opposed to SPA-ENG.

In Table \ref{tab:res_token} we show the token level results for all submitted systems. We report the F-measure for each class and the weighted average F-measure, Avg-F, which we used to rank the systems. From the results, we can see that the least difficult class to predict is the \emph{other} class, where most systems obtained an F-measure over 97\%. We can also discern that for the classes with a minority amount of tokens (\emph{ambiguous, mixed, fw, unk}) were the hardest to predict, with most systems obtaining an F-measure of 0\%. This is to be expected as we only had a small number of samples in our training and test data and in the case of the MSA-DA data set, there were no samples for \emph{fw} or \emph{unk}. Precisely because of the small amount of samples for these classes, the results do not affect in a significant way the weighed averaged F-measure score used to rank the systems. However, it is still important to correctly predict these classes in order to make a more thorough analysis of CS data.

For SPA-ENG, all the systems beat the baseline at the tweet level evaluation by at least 16\%. The best performing system here was \cite{howard:16} with an Avg-F-measure of 91.3\%, which is 1.3\% higher than the second best system \cite{hhu:16}. At the token level, all but one system outperformed the baseline. The best performing system was also \cite{howard:16} with an Avg-F-measure of 97.3\%, which is 0.4\% higher than the second best performing system (IIIT Hyderabad).\footnote{The participants did not submit a system description.}

For MSA-DA, all the submitted systems outperform the baseline at the tweet level by at least 20\%. At tweet level, \cite{hhu:16} achieved 83\%, which the highest Avg-F-measure. The second highest Avg-F-measure was achieved by \cite{gwu:16}-2. Their Avg-F-measure was 75\%. At the token level, all systems beat the baseline by at least 7\%. Also, \cite{hhu:16} succeed in achieving the highest Avg-F-measure which is 87.6\%. cite{gwu:16}-2 and \cite{gwu:16}-1 achieve the second and the third best performing systems, with Avg-F-measures of 85.1\%, 82.8\%, respectively.

It is not easy to determine overall winners because not all participants submitted a system for both language-pairs. However, for the SPA-ENG data set the system by \cite{howard:16} was the best performing at both the tweet and token level evaluations. On the other hand, the system by \cite{hhu:16} was the best performing at both tweet and token level for the MSA-DA data set.

\section{Lessons Learned}
This year we had to deal with the same issues we encountered with Twitter, including data loss and sharing restrictions. However, we decided to cope with these issues as we found it harder to identify other sources of data where we could easily search and find samples of code-switched text.

In the process of annotation, we believe that pre-annotating the data using our previous data as training helps speed up the process of in-lab annotations and thus reduces the amount of data that has to be annotated through crowd-sourcing. We took several measures to ensure we obtained high quality data from crowd-sourcing, but it still proves to be a challenge and we obtain a fair amount of noise. The problem is exacerbated in the MSA-DA set due to the fact that there is inherently considerable amount of data overlap due to homographs between the two varieties of the language. Also, a big part of the errors made by crowd-sourcing annotators involve named entities, probably because the annotators do not take the context into account in an effort to be fast and collect money quickly. 

For a future shared task, we will consider giving the crowd-sourcing annotators less choices in order to reduce error, along with providing a simpler annotation guideline with a greater amount of examples. Another thing we have in mind is to further improve our code-switched tweet evaluation by taking into account the predicted positions of the code-switch points instead of just labeling the tweets as CS or monolingual.

\section{Conclusion}
We had a very successful second shared task on language identification on code-switched data. We received submissions from 9 different teams, up from 7 teams in the previous task. Overall, this year's systems achieved a higher level of performance when compared to previous shared task. This is a good indicator of a higher understanding and interest in the problem. We also see that the results of the previous shared task influenced the decisions the participants made when designing their systems, as evidenced by the majority of the systems relying on a CRF for sequence labeling. In contrast to the previous shared task, we received submissions that used deep learning algorithms and techniques, which shows that the participants are thinking of different ways to take on the problem. On the other end, we had one rule-based system that didn't perform as well as the others and perhaps is an indicator that machine learning is definitely the baseline architecture to use for language identification  in  CS data.

In contrast to the previous shared task, the results are more consistent between token/tweet level performance, with the same teams ranking first at both levels of the same language pair. This is an indication that there were less errors made, leading to less confusion of the CS points. Also different from the previous task, this year we only looked at two different language pairs, but we maintain that these two pairs are a good representation of CS occurrences.

We have shown that there is great interest in researching language identification on code-switched data and we have provided a competitive shared task that will help push forward the development of systems and corpora with the goal of improving our understanding of code-switching.

\section*{Acknowledgments}
We would like to thank our in-lab annotators for their help in generating new test data for the shared task. We also want to thank all the participants for their contributions and interest in the problem. This work was partially supported by grants 1462143 and 1462142 from the National Science Foundation.

\bibliography{emnlp2016}
\bibliographystyle{emnlp2016}

\end{document}